\documentclass[runningheads]{llncs}
\usepackage{amsmath,amssymb} 
\usepackage{tabularx}
\usepackage{graphicx}
\usepackage{amsmath,amssymb} 
\usepackage{color}

\newcolumntype{Y}{>{\centering\arraybackslash}X}

\begin{document}
\title{StereoNet: Guided Hierarchical Refinement for Real-Time Edge-Aware Depth Prediction} 

\titlerunning{StereoNet}

\authorrunning{Khamis et al.}

\author{Sameh Khamis, Sean Fanello,  Christoph Rhemann, \\ Adarsh Kowdle, Julien Valentin, Shahram Izadi}


\institute{Google Inc.
}

\maketitle

\begin{abstract}
This paper presents StereoNet, the first end-to-end deep architecture for real-time stereo matching that runs at $60$fps on an NVidia Titan X, producing high-quality, edge-preserved, quantization-free disparity maps. A key insight of this paper is that the network achieves a sub-pixel matching precision than is a magnitude higher than those of traditional stereo matching approaches. This allows us to achieve real-time performance by using a very low resolution cost volume that encodes all the information needed to achieve high disparity precision. Spatial precision is achieved by employing a learned edge-aware upsampling function. Our model uses a Siamese network to extract features from the left and right image. A first estimate of the disparity is computed in a very low resolution cost volume, then hierarchically the model re-introduces high-frequency details through a learned upsampling function that uses compact pixel-to-pixel refinement networks. Leveraging color input as a guide, this function is capable of producing high-quality edge-aware output. We achieve compelling results on multiple benchmarks, showing how the proposed method offers extreme flexibility at an acceptable computational budget.
\end{abstract}

\section{Introduction}
Stereo matching is a classical computer vision problem that is concerned with estimating depth from two slightly displaced images. Depth estimation has recently been projected to the center stage with the rising interest in virtual and augmented reality \cite{holoportation}. It is at the heart of many tasks from 3D reconstruction to localization and tracking \cite{kinectfusion}. Its applications span otherwise disparate research and product areas including indoor mapping and architecture, autonomous cars, and human body and face tracking.

Active depth sensors like the Microsoft Kinect provide high quality depth-maps and have not only revolutionized computer vision research \cite{dou16,dou17,holoportation,fanello2013one,taylor17}, but also play an important role in consumer level applications. These active depth sensors have become very popular over the recent years with the release of many other consumer devices, such as the Intel RealSense series, the structured light sensor on iPhone X, as well as time-of-flight cameras such as Kinect V2. With the rise of Augmented Reality (AR) applications on mobile devices, there is a growing need of algorithms capable of predicting precise depth under tight computational budget. With the exception of the iPhone X, all smartphones on the market can only rely on single or dual RGB streams. The release of sparse tracking and mapping tools like ARKit and ARCore impressively demonstrate coarse and sparse geometry estimation on mobile devices. However, they lack dense depth estimation and therefore cannot enable exciting AR applications such as occlusion handling or precise interaction of virtual objects with the real world. Depth estimation using a single moving camera, akin to \cite{pradeep2013}, or dual cameras naturally became a requirement from the industry to scale AR to millions of users. 

The state of the art in passive depth relies on stereo triangulation between two (rectified) RGB images. This has historically been dominated by CRF-based approaches. These techniques obtain very good results but are computationally slow. Inference in these models amounts to solving a generally NP-hard problem, forcing practitioners in many cases to use solvers whose runtime is in the ranges of seconds \cite{meanfield} or resort to approximated solutions \cite{fanello17_hashmatch,fanello2017ultrastereo,patchCollider,sos}. Additionally, these techniques typically suffer in the presence of textureless regions, occlusions, repetitive patterns, thin-structures, and reflective surfaces. The field is slowly transitioning and since \cite{zbontar2015computing}, it started to use deep features, mostly as unary potentials, to further advance the state of the art.

Recently, deep-architectures demonstrated a high level of accuracy at predicting depth from passive stereo data \cite{mayer2016large,ilg2017flownet,kendall2017end,pang2017cascade}. Despite these significant advances, the proposed methods require vast amounts of processing power and memory. For instance, \cite{kendall2017end} have $3.5$ million parameters in their network and reach a throughput of about $0.95$ image per second on $960 \times 540$ images, and \cite{pang2017cascade} takes $0.5$ sec to produce a single disparity on a high end GPU.

In this paper we present StereoNet, a novel deep architecture that generated state of the art 720p depth maps at $60$Hz on high end GPUs. Based on our insight that deep architectures are very good to infer matches at extremely high subpixel precision we demonstrate that a very low resolution cost volume is sufficient to achieve a depth precision that is comparable to a traditional stereo matching system that operates at full resolution. To achieve spatial precision we apply edge-aware filtering stages in a multi-scale manner to deliver a high quality output. In summary the main contributions of this work are the following:

\begin{enumerate}
    \item We show that the subpixel matching precision of a deep architecture is an order of magnitude higher than those of ``traditional'' stereo approaches.
    \item We demonstrate that the high subpixel precision of the network allows to achieve the depth precision of traditional stereo matching with a very low resolution cost volume resulting in an extremely efficient algorithm.
    \item We show that previous work that introduced cost-volume in deep architectures was over-parameterized for the task and how this significantly help reducing the run-time and memory footprint of the system at little cost in accuracy.
    \item A new hierarchical depth-refinement layer that is capable of performing high-quality up-sampling that preserves edges.
    \item Finally, we demonstrate that the proposed system reaches compelling results on several benchmarks while being real-time on high end GPU architectures.
\end{enumerate}

\section{Related Work}

Depth from stereo has been studied for a long time and we refer the interested reader to \cite{scharstein2002taxonomy,hamzah2016literature} for a survey.
Correspondence search for stereo is a challenging problem and has been traditionally divided into global and local approaches.
Global approaches formulate a cost function over the image that is traditionally optimized using approaches such as Belief Propagation or Graph Cuts~\cite{besse2014pmbp,felzenszwalb2006efficient,klaus2006segment,kolmogorov2001computing}. 
Instead, local stereo matching methods (e.g. \cite{bleyer2011patchmatch}) center a support window on a pixel in the reference frame and then displace this window in the second image until the point of highest correlation is found. A major challenge for local stereo matching is to define the optimal size for the support window. On the one hand the window needs to be large to capture a sufficient amount of texture but needs to be small at the same time to avoid aggregating wrong disparity values that can lead to the well-known edge fattening effect at disparity discontinuities.
To avoid this trade-off, adaptive support approaches weigh the influence of each pixel inside the support region based on e.g. its color similarity to the central pixel.

Interestingly adaptive support weight approaches were cast as cost volume filtering in~\cite{hosni2013fast}: a three-dimensional cost volume is constructed by computing the per-pixel matching costs at all possible disparity levels. This cost volume is then filtered with a weighted average filter. This filtering propagates local information in the spatial and depth domains producing a depth map that preserves edges across object discontinuities.

For triangulation based stereo matching system the accuracy of depth is directly linked to the precision to which the corresponding pixel in the other image can be located. Therefore, previous work strives to do matching with sub-pixel precision. The complexity of most algorithms scale linearly with the number of disparities evaluated so while one approach is to build a large cost volume with very fine grained disparity steps this is computationally in-feasible. Many algorithms therefore start with discrete matching and then refine these matches by fitting a local curve such as a parabolic fit to the cost function between the discrete disparity candidates (see e.g. \cite{yang2007,Nehab2005}). Other works are based on continuous optimization strategies \cite{Ranftl2012} or on phase
correlation \cite{Sanger88}. It was shown in \cite{pinggera2014} that under realistic conditions the bound for subpixel precision is $1/10th$ of a pixel while the theoretical limit under noise free conditions was found to be 10 times lower \cite{Delon2007}. We demonstrate that this traditional wisdom does not hold true for learning-based approaches and we can achieve a subpixel precision of $1/30th$ of a pixel.

Recent work has progressed to using end-to-end learning for stereo matching. Various approaches combined a learned patch embedding or matching cost with global optimization approaches like semiglobal matching (SGM) for refinement~\cite{zagoruyko2015learning}. \cite{chen2015deep} learn a multi-scale embedding model followed by an MRF. \cite{zbontar2016stereo,zbontar2015computing} learn to match image patches followed by SGM. \cite{luo2016efficient} learn to match patches using a Siamese feature network and optimize globally with SGM as well. ~\cite{shaked2017improved} uses a multi-stage approach where a highway network architecture is first used to compute the matching costs and then another network is used in postprocessing to aggregate and pool costs.

Other works attempted to solve the stereo matching problem end-to-end without postprocessing. 
\cite{mayer2016large,ilg2017flownet} train end-to-end an encoder-decoder network for disparity and flow estimation achieving state-of-the-art results on existing and new benchmarks. Other end-to-end approaches used multiple refinement stages that converge to the right disparity hypotheses. \cite{gidaris2017detect} proposed a generic architecture for labeling problems, including depth estimation, that is trained end-to-end to predict and refine the output. \cite{pang2017cascade} proposed a cascaded approach to refine predicted depth iteratively. Iterative refinement approaches, while showing good performance on various benchmarks, tend to require a considerable amount of computational resources.

More closely related to our work is \cite{kendall2017end} who used the concept of cost volume filtering but trained both the features and the filters end-to-end achieving impressive results. DeepStereo~\cite{flynn2016deepstereo} used a plane-sweep volume to synthesize novel views from multi-view stereo input. Contrary to prior work, we are interested in an end-to-end learning stereo pipeline that can run in real-time, therefore we start from a very low resolution cost volume, which is then upsampled with learned, edge aware filters.

\section{StereoNet algorithm}
\subsection{Preliminaries}

Given pairs of input images we aim to train an end-to-end disparity prediction pipeline. One approach to train such pipeline is to leverage a generic encoder-decoder network. An encoder distills the input through a series of contracting layers to a bottleneck that captures the details most relevant to the task in training, and the decoder reconstructs the output from the representation captured in the bottleneck layer through a series of expanding layers. While this approach is widely successful across various problems, including depth prediction\cite{mayer2016large,ilg2017flownet,pang2017cascade}, they lack several qualities we care about in stereo algorithm.

First of all, this approach does not capture any geometric intuition about the stereo matching problem. Stereo prediction is first-and-foremost a correspondence matching problem, so we aimed to design an algorithm that can be adapted without retraining to different stereo cameras with varying resolutions and baselines. Secondly, we note that similar approaches are evidently overparameterized for problems where the prediction is a pixel-to-pixel mapping that does not involve any warping of the input, and thus likely to overfit.

\begin{figure}[t]
\centering
\includegraphics[width=0.9\columnwidth]{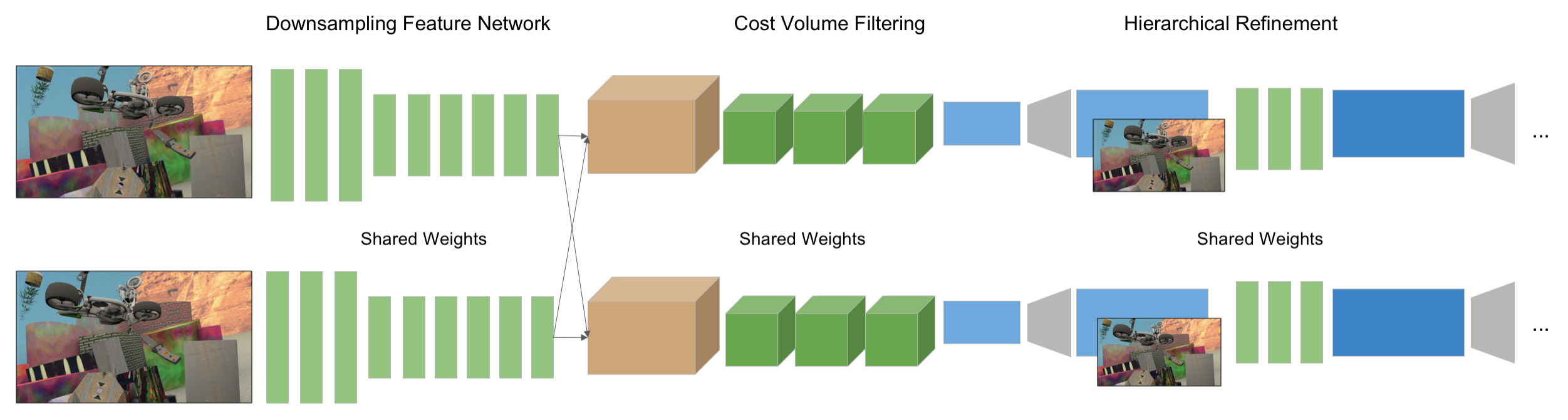}
\caption{Model architecture. A two stage approach is proposed: first we extract image features at a lower resolution using a Siamese network. We then build a cost volume at that resolution by matching the features along the scanlines, giving us a coarse disparity estimate. We finally refine the results hierarchically to recover small details and thin structures.}
\label{fig:architecture}
\end{figure}

Our approach to stereo matching incorporates a design that leverages the problem structure and classical approaches to tackle it, akin to~\cite{kendall2017end}, while producing edge-preserving output using compact context-aware pixel-to-pixel refinement networks. An overview of the architecture of our model is illustrated in Figure~\ref{fig:architecture} and detailed in the following sections.

\subsection{Coarse Prediction: Cost Volume Filtering}
\label{sec:costvolume}
Stereo system are in general solving a correspondence problem. The problem classically boils down to forming a disparity map by finding a pixel-to-pixel match between two rectified images along their scanlines. The desire for a smooth and edge-preserving solution led to approaches like cost volume filtering~\cite{hosni2013fast}, which explicitly model the matching problem by forming and processing a 3D volume that jointly solves across all candidate disparities at each pixel. While \cite{hosni2013fast} directly used color values for the matching, we compute a feature representation at each pixel that is used for matching.

\subsubsection{Feature Network} The first step of the pipeline finds a meaningful representation of image patches that can be accurately matched in the later stages. We recall that stereo suffer from textureless regions and traditional methods solve this issue by aggregating the cost using large windows. We replicate the same behavior in the network by making sure the features are extracted from a big receptive field. In particular, we use a feature network with shared weights between the two input images (also known as a Siamese network). We first aggressively downsample the input images using K $5\times5$ convolutions with a stride of 2, keeping the number of channels at 32 throughout the downsampling. In our experiments we set K to 3 or 4. We then apply 6 residual blocks~\cite{he2016deep} that employ $3\times3$ convolutions, batch-normalization~\cite{ioffe2015batch}, and leaky ReLu activations ($\alpha=0.2$)~\cite{maas2013rectifier}. Finally, this is processed using a final layer with a $3\times3$ convolution that does not use batch-normalization or activation. The output is a 32-dimensional feature vector at each pixel in the downsampled image. This low resolution representation is important for two reasons: 1) it has a big receptive field, useful for textureless regions. 2) It keeps the feature vectors compact.

\subsubsection{Cost Volume} At this point, we form a cost volume at the coarse resolution by taking the difference between the feature vector of a pixel and the feature vectors of the matching candidates. We noted that asymmetric representations in general performed well, and concatenating the two vectors achieved similar results in our experiments.

At this stage, a traditional stereo method would use a winner-takes-all (WTA) approach that picks the disparity with the lowest Euclidean distance between the two feature vectors. Instead, here we let the network to learn the right metric by running multiple convolutions followed by non-linearities.

In particular, to aggregate context across the spatial domain as well as the disparity domain, we filter the cost volume with four 3D convolutions with a filter size of $3 \times 3 \times 3$, batch-normalization, and leaky ReLu activations. A final $3 \times 3 \times 3$ convolutional layer that does not use batch-normalization or activation is then applied, and the filtering layers produce a 1-dimensional output at each pixel and candidate disparity.

For an input image of size $W \times H$ and evaluating a maximum of $D$ candidate disparities, our cost volume is of size $W / 2^K \times H / 2^K \times (D + 1) / 2^K$ for $K$ downsampling layers. In our design of StereoNet we targeted a compact approach with a small memory footprint that can be potentially deployed to mobile platforms. Unlike~\cite{kendall2017end} who form a feature representation at quarter resolution and aggregate cost volumes across multiple levels, we note that most of the time and compute is spent matching at higher resolutions, while most of the performance gain comes from matching at lower resolutions. We validate this claim in our experiments and show that the performance loss is not significant in light of the speed gain.
The reason for this is that the network achieves a magnitude higher sub-pixel precision than traditional stereo matching approaches. Therefore, matching at higher resolutions is not needed.

\subsubsection{Differentiable $\arg\min$}
We typically would select the disparity with the minimum cost at each pixel in the filtered cost volume using $\arg\min$. For a pixel $i$ and a cost function over disparity values $C(d)$, the selected disparity value $d_i$ is defined as:

\begin{equation}
d_i = \arg\min_d C_i(d).
\end{equation}

\noindent
This however fails to learn since $\arg\min$ is a non-differentiable function. We considered two differentiable variants in our approach. The first of which is soft $\arg\min$, which was originally proposed in~\cite{chapelle2010gradient} and was used in~\cite{kendall2017end}. Effectively, the selected disparity is a softmax-weighted combination of all the disparity values:

\begin{equation}
d_i = \sum_{d=1}^{D} d \cdot \frac{\exp(-C_i(d))}{\sum_{d'} \exp(-C_i(d')}.
\end{equation}

\noindent
The second differentiable variant is a probabilistic selection that samples from the softmax distribution over the costs:

\begin{equation}
d_i = d, \text{ where } d \sim \frac{\exp(-C_{i}(d))}{\sum_{d'} \exp(-C_{i}(d')}.
\end{equation}

\noindent
Differentiating through the sampling process uses gradient estimation techniques to learn the distribution of disparities by minimizing the expected loss of the stochastic process. While this technique has roots in policy gradient approaches in reinforcement learning~\cite{williams1992simple}, it was recently formulated as stochastic computation graphs in~\cite{schulman2015gradient} and applied to RANSAC-based camera localization in~\cite{brachmann2017dsac}.  Additionally, the parallel between the two differentiable variants we discussed is akin to that between soft and hard attention networks~\cite{xu2015show}.

Unfortunately the probabilistic approach significantly underperformed in our experiments, even with various variance reduction techniques~\cite{xu2015show}. We expect that this is because it preserves hard selections. This trait is arguably critical in many applications, but in our model it is superseded by the ability of soft $\arg\min$ to regress subpixel-accurate values. This conclusion is supported by the literature on continuous action spaces in reinforcement learning~\cite{lillicrap2015continuous}. The soft $\arg\min$ selection was consequently faster to converge and easier to optimize, and it is what we chose to use in our experiments.

\subsection{Hierarchical Refinement: Edge-Aware Upsampling}

The downside to relying on coarse matching is that the resulting myopic output lacks fine details. To maintain our compact design, we approach this problem by learning an edge-preserving refinement network. We note that the network's job at this stage is to dilate or erode the disparity values to blend in high-frequency details using the color input as guide, so a compact network that learns a pixel-to-pixel mapping, similar to networks employed in recent computational photography work~\cite{chen2017fast,chen2017photographic,gharbi2017deep}, is an appropriate approach. Specifically, we task the refinement network of only finding a residual (or a delta disparity) to add or subtract from the coarse prediction.

\begin{figure}[t]
\centering
\includegraphics[width=0.9\columnwidth]{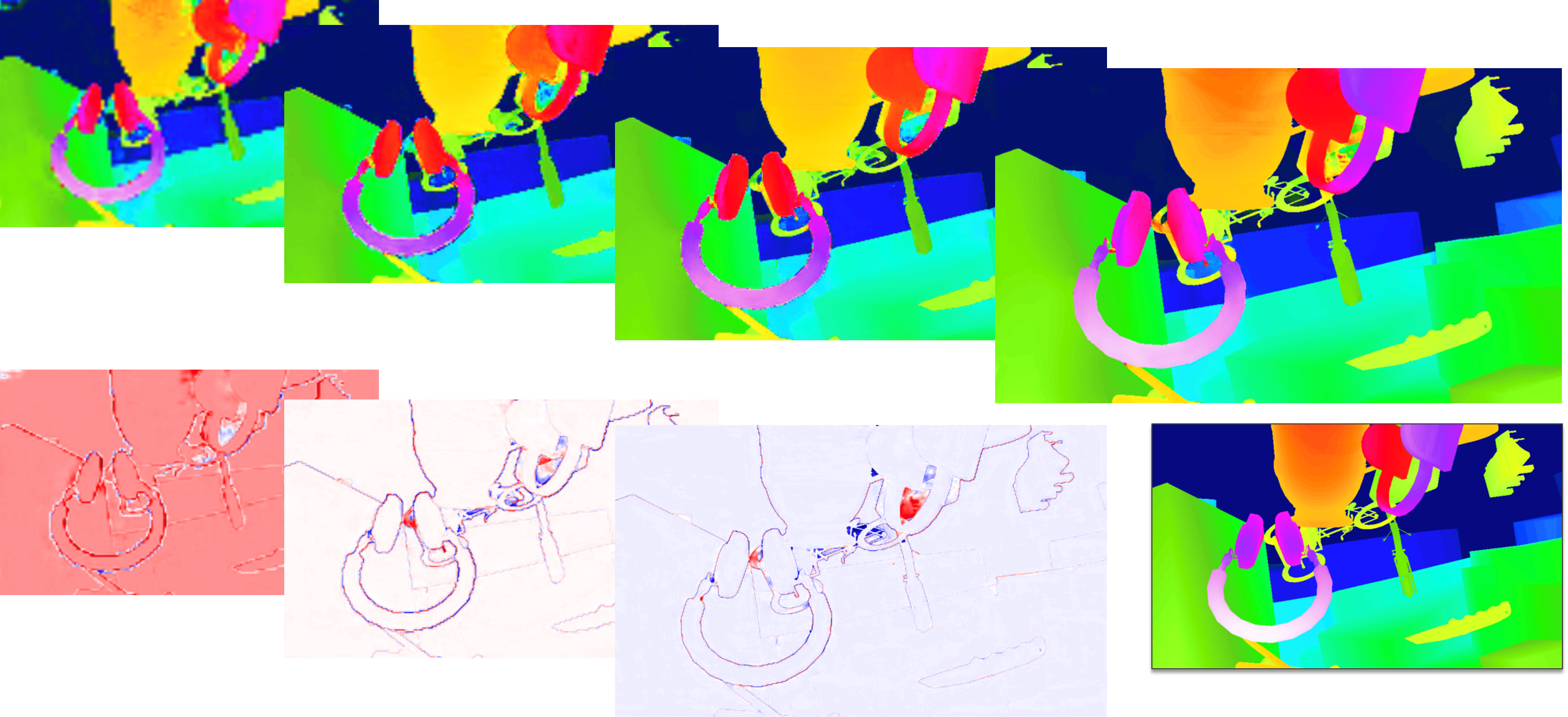}
\caption{Hierarchical refinement results. The result at each stage (top row), starting with the cost volume output in the top left corner, is updated with the output of the corresponding refinement network (bottom row). The refinement network output expectedly dilates and erodes around the edges using the color input as guide. The groundtruth is shown in the lower right corner. The average endpoint error at each stage for this example is: 3.27, 2.34, 1.80, and 1.26 respectively. Zoom in for details.}
\label{fig:refinement}
\end{figure}

Our refinement network takes as input the disparity bilinearly upsampled to the output size as well as the color resized to the same dimensions. Recently deconvolutions were shown to produce checkerboard artifacts, so we opted to use bilinear upsampling and convolutions instead~\cite{odena2016deconvolution}.  The concatenated color and disparity first pass through a $3 \times 3$ convolutional layer that outputs a 32-dimensional representation. This is then passed through 6 residual blocks that, again, employ $3\times3$ convolutions, batch-normalization, and leaky ReLu activations ($\alpha=0.2$). We use atrous convolutions in these blocks to sample from a larger context without increasing the network size~\cite{papandreou2015modeling}. We set the dilation factors for the residual blocks to 1, 2, 4, 8, 1, and 1 respectively. This output is then processed using a $3\times3$ convolutional layer that does not use batch-normalization or activation. The output of this network is a 1-dimensional disparity residual that is then added to the previous prediction. We apply a ReLu to the sum to constrain disparities to be positive.

In our experiments we evaluated hierarchically refining the output with a cascade of the described network, as well as applying a single refinement that upsamples the coarse output to the full resolution in one-shot. Figure~\ref{fig:refinement} illustrates the output of the refinement layer at each level of the hierarchy as well as the residuals added at each level to recover the high-frequency details. The behavior of this network is reminiscent of joint bilateral upsampling~\cite{kopf2007joint}, and indeed we believe this network is a learned edge-aware upsampling function that leverages a guide image.

\subsection{Loss Function}
We train StereoNet in a fully supervised manner using groundtruth-labeled stereo data. We minimize the hierarchical loss function:

\begin{equation}
L = \sum_k \rho(d_i^k - \hat{d}_i),
\end{equation}

\noindent
where $d_i^k$ is the predicted disparity at pixel $i$ at the $k$-th refinement level, with $k=0$ denoting the output pre-refinement, and $\hat{d}_i$ is the groundtruth disparity at the same pixel. The predicted disparity map is always bilinearly upsampled to match the groundtruth resolution. Finally, $\rho(.)$ is the two-parameter robust function from~\cite{barron2017more} with its parameters set as $\alpha=1$ and $c=2$, approximating a smoothed L1 loss.

\subsection{Implementation details}

We implemented and trained StereoNet using Tensorflow~\cite{abadi2016tensorflow}. All our experiments were optimized using RMSProp~\cite{hinton2012neural} with an exponentially-decaying learning rate initially set to $1\mathrm{e}{-3}$. Input data is first normalized to the range $[-1, 1]$. We use a batch size of 1 and we do not crop because of the smaller model size, unlike~\cite{kendall2017end}.

Our network needs around $150k$ iterations to reach convergence. We found that, intuitively, training with the left and right disparity maps for an image pair at the same time significantly sped up the training time. On smaller datasets where training from scratch would be futile, we fine-tuned the pre-trained model for an additional $50k$ iterations.

\section{Experiments}
Here, we evaluate our system on several datasets and demonstrate that we achieve high quality results at a fraction of the computational cost required by the state of the art.

\subsection{Datasets and Setup}
We evaluated StereoNet quantitatively and qualitatively on three datasets: Scene Flow~\cite{mayer2016large},  KITTI 2012 ~\cite{geiger2012we} and KITTI 2015 ~\cite{Menze2015CVPR}. Scene Flow is a large synthetic stereo dataset suitable for deep learning models. However, the other two KITTI datasets, while more comparable to a real-world setting, are too small for full end-to-end training. We followed previous end-to-end approaches by initially training on Scene Flow and then individually fine-tuning the resulting model on the KITTI datasets~\cite{kendall2017end,pang2017cascade}. Finally, we compare against prominent state-of-the-art methods in terms of both accuracy and runtime to show the viability of our approach in real-time scenarios.

Additionally, we performed an ablation study on the Scene Flow dataset using four variants of our model. We evaluated setting the number of downsampling convolutions $K$ (detailed in Section~\ref{sec:costvolume}) to $3$ and $4$. This controls the resolution at which the cost volume is formed. The cost volume filtering is exponentially faster with more aggressive downsampling, but comes at the expense of increasingly losing details around thin structures and small objects. The refinement layer can bring in a lot of the fine details, but if the signal is completely missing from the cost volume, it is unlikely to recover them. Additionally we evaluated using $K$ refinement layers to hierarchically recover the details at the different scales versus using a single refinement layer to upsample the cost volume output directly to the desired final resolution.

\subsection{Subpixel Precision}
The precision of a depth system is usually a crucial variable when choosing the right technology for a given application. A triangulation system with a baseline $b$, a focal length $f$ and a subpixel precision $\delta$ has an error $\epsilon$ which increases quadratically with the distance $Z$: $\epsilon = \frac{\delta Z^2}{bf}$ \cite{Szeliski_book}. Competitive technologies such as Time-of-Flight do not suffer from this issue, which makes them appealing for long range applications such as room scanning and reconstruction. Despite this it has been demonstrated that multipath effects in ToF systems can distort geometry even in close-up tasks such as object scanning \cite{hyperdepth}. Long range precision remains as one of the main arguments against a stereo system and in favor of ToF.

Here we show that deep architectures are a breakthrough in terms of subpixel precision and therefore they can compete with other technologies not only for short distances but as well as in long ranges. Traditional stereo matching methods perform a discrete search and then a parabola interpolation to retrieve the accurate disparity. This methods usually leads to a subpixel precision $\sim 0.25$ pixels, that roughly correspond to $4.5$ cm error at $3$m distance for a system with a $55$ cm baseline such as the Intel Realsense D415.
\begin{figure}[t]
\centering
\includegraphics[width=\columnwidth]{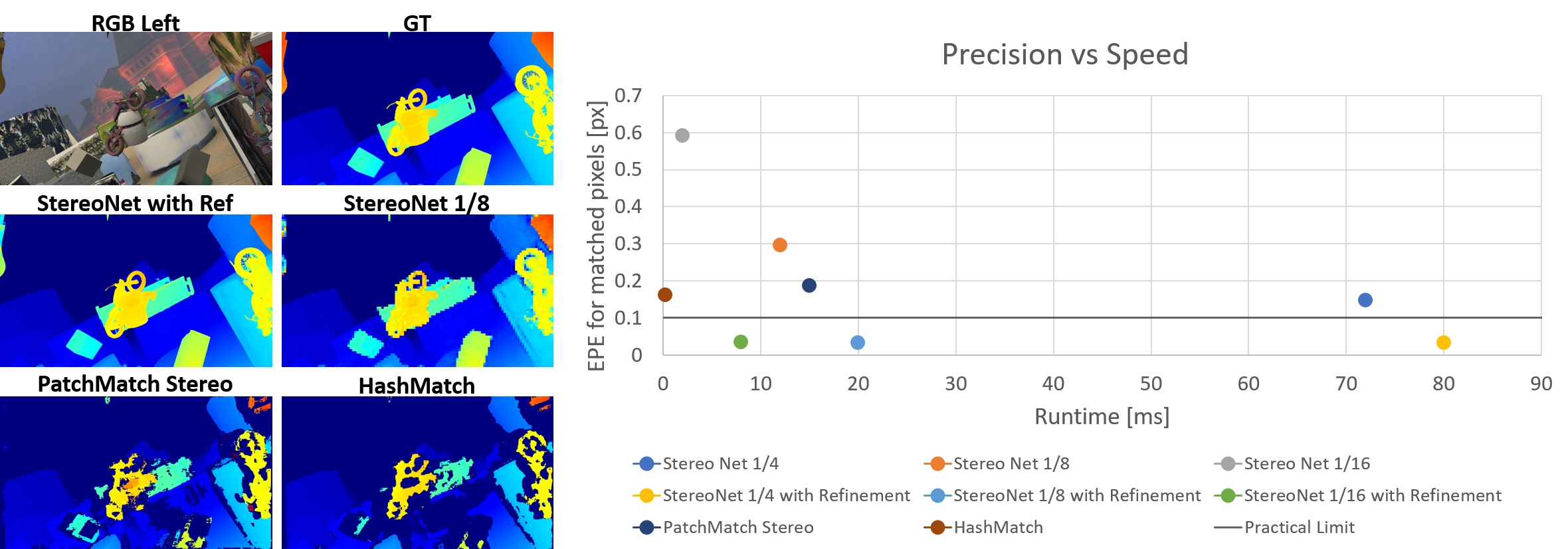}
\caption{Subpixel precision in stereo matching. We demonstrate that StereoNet achieves a subpixel precision of $0.03$, which is one order of magnitude lower than traditional stereo approaches. The lower bound of traditional approaches was found to be $1/10th$ under realistic conditions (see \cite{pinggera2014}) which we indicate by the black line. Moreover, our method can run in real-time on 720p images.}
\label{fig:cost_volume_exp}
\end{figure}

To assess the precision of our method, we used the evaluation set of Scene Flow and we computed the average error only for those pixels that were correctly matched at integer locations. Results correspond to the average of over a hundred million pixels and are reported in Figure~\ref{fig:cost_volume_exp}. From this figure, it is important to note that: (1) the proposed method achieves a subpixel precision of $\mathbf{0.03}$ which is one order of magnitude lower than traditional stereo matching approaches such as \cite{bleyer2011patchmatch,fanello17_hashmatch,fanello2017ultrastereo}; 
(2) the refinement layers are performing very similarly irrespective of the resolution of the cost volume;
(3) without any refinement the downsampled cost volume can still achieve a subpixel precision of $0.03$ in the low resolution output. However, the error increases, almost linearly, with the downsampling factor.

Note that a subpixel precision of $0.03$ means that the expected error is less than $5$mm at $3$m distance from the camera (Intel Realsense D415). This result makes triangulation systems very appealing and comparable with ToF technology without suffering from multi-path effects.

\subsection{Quantitative Results}
We now evaluate the model on standard benchmarks proving the effectiveness of the proposed methods and the different trade-offs between the resolution of the cost volume and the precision obtained.

\paragraph{\textbf{SceneFlow.}} Although this data is synthetically generated, the evaluation sequences are very challenges due to the presence of occlusions, thin structures and large disparities. We evaluated our model reporting the end point error (EPE) in Table \ref{tab:flyingthings3d}. 

\begin{figure}[t]
\centering
\includegraphics[width=0.9\columnwidth]{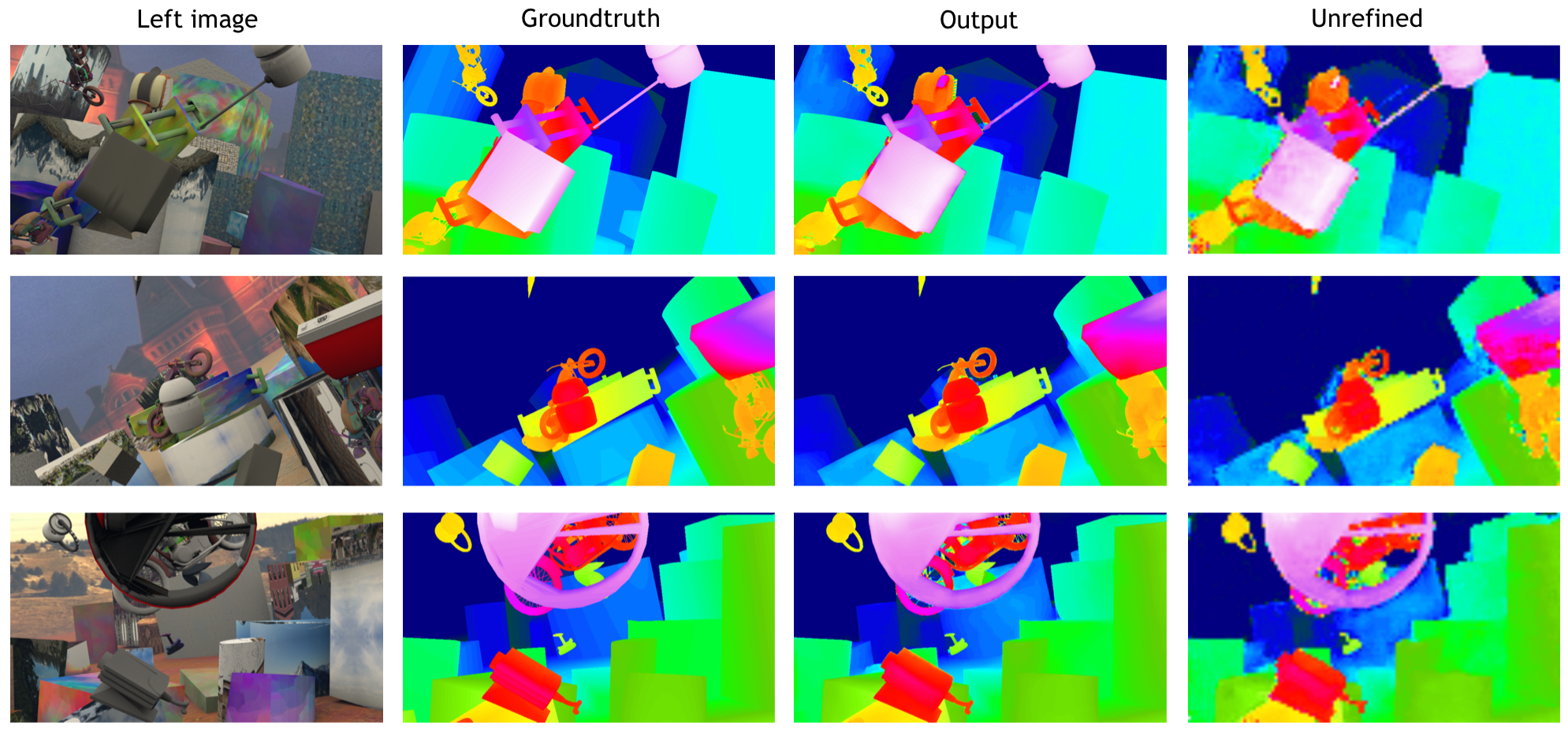}
\caption{Qualitative results on the FlyingThings3D test set. The proposed two-stage architecture is able to recover very fine details despite the low resolution at which we form the cost volume. }
\label{fig:results}
\end{figure}

A single, unrefined model, i.e. using only the cost volume output at 1/8 of the resolution, achieves an EPE of $2.48$ which is better than the full model presented in \cite{kendall2017end}, which reaches an EPE of $2.51$. Notice that our unrefined model is composed of $~360k$ parameters and runs at $12$ msec at the $960 \times 540$ input resolution, whereas \cite{kendall2017end} uses 3.5 million parameter with a runtime of $950$ msec on the same resolution. Our best, multi-scale architecture achieves the state-of-the-art error of $1.1$, which is also lower than the one reported in very recent methods such as~\cite{pang2017cascade}. Qualitative examples can be found in Figure~\ref{fig:results}. Notice how the method recovers very challenging fine details.

\begin{figure}[t]
\centering
\includegraphics[width=0.9\columnwidth]{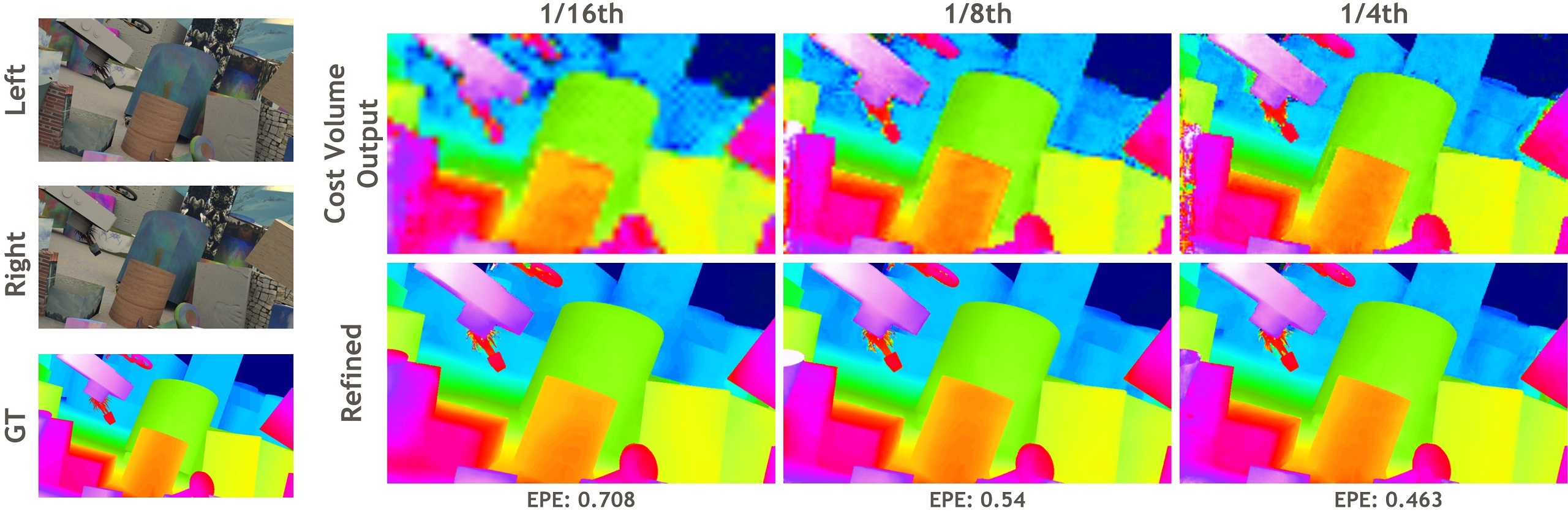}
\caption{Cost volume comparisons. A cost volume at 1/16 resolution has already the information required to produce high quality disparity maps. This is evident in that post refinement we recover challenging thin structures and the overall end point error (EPE) is below one pixel.}
\label{fig:cost_volume_qual}
\end{figure}

One last consideration regards the resolution of the cost volume. On one hand we proved that a coarse cost volume already carries all the information needed to retrieve a very high subpixel precision, i.e. high disparity resolution. On the other hand, downsampling the image may lead to a loss in spatial resolution, therefore thin structures cannot be reconstructed if the output of the cost volume is very coarse. Here we demonstrate that a volume at 1/16 of the resolution is powerful enough to recover very challenging small objects. Indeed in Figure ~\ref{fig:cost_volume_qual}, we compare the output of the three cost volumes at 1/4, 1/8, 1/16 resolutions where we also applied the refinement layers. We can observe that the fine structures that are missed in the 1/16 resolution disparity map are correctly recovered by the upsampling strategy we propose. The cost volume at 1/4 is not necessary to achieve a compelling results and this is an important finding for mobile applications. As showed in the previous subsection, even at low resolution the network achieves a subpixel precision of $1/30$th pixel. However, we want to also highlight that to achieve state of the art precision on multiple benchmarks, the cost volume resolution becomes an important factor as demonstrated in Table~\ref{tab:flyingthings3d}.

\begin{table}[t]
    \centering
    \begin{tabularx}{\textwidth}{|X|YYYY|}
    \hline
    & EPE all & EPE nocc & EPE all, unref & EPE nocc, unref\\
    \hline
    8x, multi & $\mathbf{1.101}$ & 0.768 & 2.512 & 1.795 \\
    8x, single & 1.532 & 1.058 & 2.486 & 1.784 \\
    16x, multi & 1.525 & 1.140 & 3.764 & 2.912 \\
    16x, single & 1.974 & 1.476 & 3.558 & 2.773 \\
    \hline
    \scriptsize{CG-Net Fast} \cite{kendall2017end} & 7.27  & - & - & - \\
    \scriptsize{CG-Net Full} \cite{kendall2017end} & 2.51 & - & - & - \\
     CRL \cite{pang2017cascade} & 1.32 & - & - & - \\ 
    \hline
    \end{tabularx}
    \caption{Quantitative evaluation on SceneFlow. We achieve state of the art results compared to recent deep learning methods. We compare four variants of our model which vary in the resolution at which the cost volume is formed (8x vs 16x) and the number of refinement layers (multiple vs single).}
\label{tab:flyingthings3d}
\end{table}

\begin{figure}[t]
\centering
\includegraphics[width=0.9\linewidth]{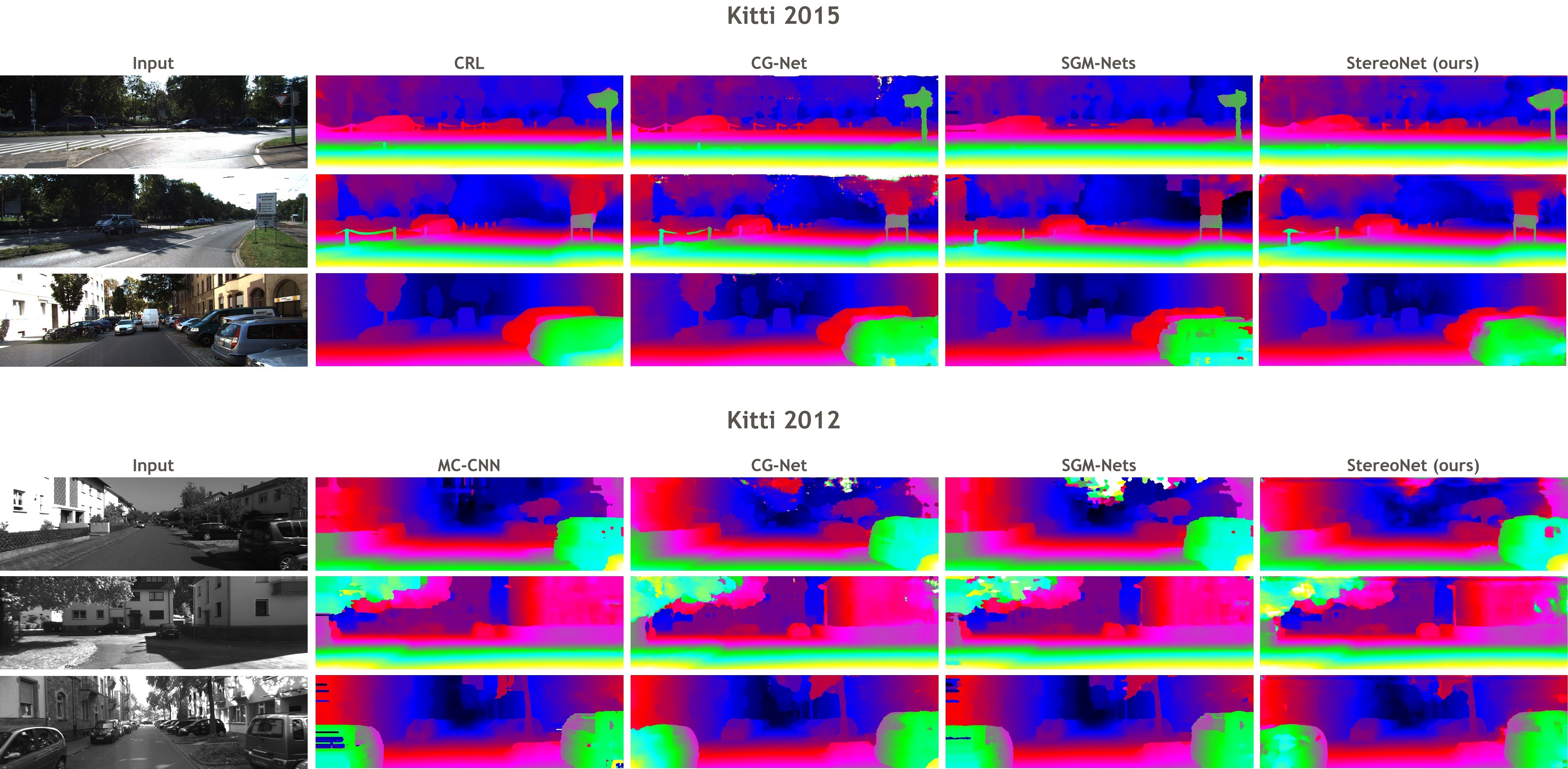}
\caption{Qualitative Results on Kitti 2012 and Kitti 2015. Notice how our method preserves edge and recovers details compared to the fast \cite{sgm-net}. State of the art methods are one order of magnitude slower than the proposed approach.}
\label{fig:kitti12_15}
\end{figure}

\begin{table}[t]
    \centering
    \begin{tabularx}{1.0\textwidth}{|X|YYYYY|}
    \hline
    & Out-Noc & Out-All & Avg-Noc & Avg-All & Runtime\\
    \hline
    \scriptsize{StereoNet} & 4.91 & 6.02 & 0.8 & 0.9 & 0.015s\\
    \hline
    \scriptsize{CG-Net} \cite{kendall2017end} & 2.71 & 3.46 & 0.6 & 0.7 & 0.9s \\
    \scriptsize{MC-CNN \cite{zbontar2016stereo}}  & 3.9 & 5.45 & 0.7  & 0.9 & 67s\\ 
    \scriptsize{SGM-Net \cite{sgm-net}}  & 3.6 & 5.15 & 0.7  & 0.9 & 67s\\ 
    \hline
    \end{tabularx}
    \caption{Quantitative evaluation on Kitti 2012. For StereoNet we used a model with a downsampling factor of $8$ and $3$ refinement levels. We report the percentage of pixels with error bigger than $2$, as well as the overall EPE in both non occluded (Noc) and all the pixels (All).}
\label{tab:kitti12}
\end{table}

\paragraph{\textbf{Kitti.}} Kitti is a prominent stereo benchmark that was captured by driving a car equipped with cameras and a laser scanner~\cite{geiger2012we}. The dataset is very challenging due to the huge variability, reflections, overexposed areas and more importantly, the lack of a big training set. Despite this, we provide the results on Kitti 2012 in Table \ref{tab:kitti12}. Our model uses a downsampling factor of $8$ for the cost volume and $3$ refinement steps. Among the top-performing methods, we compare to three significant ones. Current state of the art \cite{kendall2017end}, achieves an EPE of $0.6$, but it has a running time of $0.9$ seconds per image and uses a multi-scale cost volume and several 3D deconvolutions. The earlier deep learning-based stereo matching approach of \cite{zbontar2016stereo} takes $67$ seconds per image and has higher error ($0.9$) compared to our method that runs at $0.015$s per stereo pair. The SGM-net \cite{sgm-net} has an error comparable to ours. Although we do not reach state of the art results, we believe that the produced disparity maps are very compelling as shown in Figure ~\ref{fig:kitti12_15}, bottom. We analyzed the source of errors in our model and we found that most of the wrong estimates are around reflections, which result in a wrong disparity prediction, as well as occluded regions, which do not have a correspondence in the other view. These areas cannot be explained by the data and the problem can then be formulated as an inpainting task, which our model is not trained for. State of the art \cite{pang2017cascade} uses a hour-glass like architecture in their refinement step, that has been shown to be really effective for inpainting purposes \cite{hourglass-inpainting}. This is certainly a valid solution to handle those invalid areas, however it requires significant additional computational resources.  We believe that the simplicity of the proposed architecture shows important insights and it can lead the way to interesting directions to overcome the current limitations.

Similarly, we evaluated our algorithm on Kitti 2015 and report the results in Tab. \ref{tab:kitti15}, where similar considerations can be made. In Figure ~\ref{fig:kitti12_15} top, we show some examples from the test data.

\begin{table}[t]
    \centering
    \begin{tabularx}{0.9\textwidth}{|X|YYYY|}
    \hline
    & D1-bg & D1-fg & D1-all & Runtime\\
    \hline
    \scriptsize{StereoNet} & 4.30 & 7.45 & 4.83& 0.015s\\
    \hline
    \scriptsize{CRL \cite{pang2017cascade}}  & 2.48 & 3.59 & 2.67 & 0.5s \\
    \scriptsize{CG-Net Full} \cite{kendall2017end} & 2.21 & 6.16 & 2.87 & 0.9s \\
    \scriptsize{MC-CNN \cite{zbontar2016stereo}}  & 2.89 & 8.88 & 3.89 & 67s\\ 
    \scriptsize{SGM-Net \cite{sgm-net}}  & 2.66 & 8.64 & 3.66  & 67s\\ 
    \hline
    \end{tabularx}
    \caption{Quantitative evaluation on Kitti 2015. For StereoNet we used a model with a downsampling factor of $8$ and $3$ refinement levels. We report the percentage of pixels with error bigger than $1$ in background regions (bg), foreground areas (fg), and all.}
\label{tab:kitti15}
\end{table}


\begin{figure}[t]
\centering
\includegraphics[width=0.9\columnwidth]{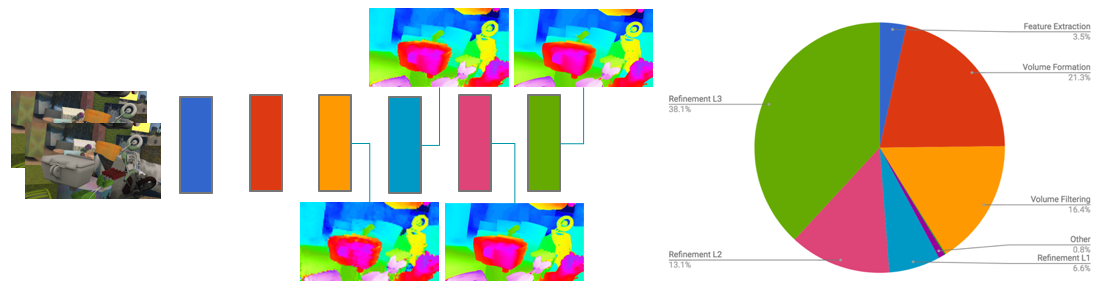}
\caption{Runtime analysis of StereoNet. Breakdown of the running time. Notice how most of the time is spent at the last level of refinement. }
\label{fig:breakdown}
\end{figure}

\subsection{Running Time Analysis}
We conclude this section with a breakdown of the running time of our algorithm. Readers interested in real-time applications would find useful to understand where the bottlenecks are. The current algorithm runs at $60$fps on an NVidia Titan X and in Fig. \ref{fig:breakdown} of the whole running time. Notice how feature extraction, volume formation and filtering take less than half of the whole computation ($41\%$), and the most time consuming steps are the refinement stage: the last level of refinement done at full resolution is using $38 \%$ of the computation.

\section{Discussion}
We presented StereoNet, the first real-time, high quality end-to-end architecture for passive stereo matching. We started from the insight that a low resolution cost volume contains most of the information to generate high-precision disparity maps and to recover thin structures given enough training data. We demonstrated a subpixel precision of $1/30th$ pixel, surpassing limits published in the literature. Our refinement approach hierarchically recovers high-frequency details using the color input as guide, drawing parallels to a data-driven joint bilateral upsampling operator. The main limitation of our approach is due to the lack of supervised training data: indeed we showed that when enough examples are available, our method reaches state of the art results. To mitigate this effect, our future work involves a combination of supervised and self-supervised learning \cite{activestereonet} to augment the training set.

\bibliographystyle{splncs04}
\bibliography{paper}
\end{document}